\documentclass[10pt,twocolumn,letterpaper]{article}

\usepackage[pagenumbers]{cvpr}

\usepackage{url}
\usepackage{booktabs}
\usepackage{amsfonts}
\usepackage{nicefrac}
\usepackage{microtype}
\usepackage{lipsum}
\usepackage{fancyhdr}
\usepackage{graphicx}
\usepackage{subcaption}
\usepackage{wrapfig}
\usepackage{caption}
\usepackage{float}

\usepackage{multirow}
\usepackage{threeparttable}

\usepackage[table]{xcolor}
\usepackage{colortbl}
\usepackage{color,soul}
\sethlcolor{yellow}
\definecolor{mygreen}{RGB}{130,179,102}
\definecolor{myblue}{RGB}{108,142,191}
\definecolor{myred}{RGB}{191,108,108}
\definecolor{darkgreen}{RGB}{0,190,0}

\usepackage{amsmath}
\usepackage{amssymb}
\usepackage{bm}
\usepackage{dsfont}

\usepackage{algorithm}
\usepackage{algorithmic}
\usepackage{listings}
\usepackage{verbatim}
\usepackage{comment}

\usepackage{pifont}

\usepackage{textcomp}
\usepackage{relsize}

\usepackage{cancel}

\usepackage{tikz}
\usetikzlibrary{decorations.pathreplacing, positioning}

\usepackage{appendix}
\usepackage{afterpage}

\usepackage{etoolbox}
\makeatletter
\AfterEndEnvironment{algorithm}{\let\@algcomment\relax}
\AtEndEnvironment{algorithm}{\kern2pt\hrule\relax\vskip3pt\@algcomment}
\let\@algcomment\relax
\newcommand\algcomment[1]{\def\@algcomment{\footnotesize#1}}
\renewcommand\fs@ruled{\def\@fs@cfont{\bfseries}\let\@fs@capt\floatc@ruled
  \def\@fs@pre{\hrule height.8pt depth0pt \kern2pt}%
  \def\@fs@post{}%
  \def\@fs@mid{\kern2pt\hrule\kern2pt}%
  \let\@fs@iftopcapt\iftrue}
\makeatother

\definecolor{pearFour}{HTML}{588F27}
\definecolor{pearFith}{HTML}{ECF0F1}
\definecolor{pearDark}{HTML}{2980B9}
\definecolor{pearDarker}{HTML}{1D2DEC}
\definecolor{pearLight}{HTML}{F7E0D5}
\definecolor{cvprblue}{rgb}{0.21,0.49,0.74}
\usepackage[pagebackref,breaklinks,colorlinks,allcolors=cvprblue]{hyperref}

\def\paperID{*****}
\def\confName{CVPR}
\def\confYear{2026}

\title{
\textit{SynCo}: Synthetic Hard Negatives for Contrastive Visual Representation Learning
}

\author{
Nikos Giakoumoglou \qquad Tania Stathaki\\
Imperial College London\\
{\tt\small \{nikos,tania\}@imperial.ac.uk}\vspace{.5em}\\
{Code: \url{https://github.com/giakoumoglou/synco}}\\
}

\begin{document}
\maketitle

%%%%%%%%% MAIN PAPER
\begin{abstract}
Contrastive learning relies on informative negatives to shape the representation space, yet obtaining hard negatives is costly, often requiring large batch sizes or extensive memory banks. We propose SynCo (\underline{\textbf{{Sy}}}nthetic \underline{\textbf{{n}}}egatives in \underline{\textbf{{Co}}}ntrastive learning), an approach that synthesizes hard negatives directly in the representation space from cached queue embeddings, with no additional forward passes or input-space processing. We find that six lightweight synthesis strategies, exhaustively covering the geometric, stochastic, and adversarial perturbation families, consistently improve learned representations at negligible computational cost. Although applicable to any InfoNCE-based contrastive objective, we demonstrate SynCo within the MoCo framework. On ImageNet ILSVRC-2012 linear evaluation at 200 epochs, SynCo yields improvements of \textbf{+0.4\%} over MoCo-v2 and \textbf{+1.0\%} over MoCHi. Unlike MoCHi, which degrades at extended pretraining schedules (underperforming MoCo-v2 by 2.4\% at 800 epochs), SynCo does not: with a simple synthetic negative schedule, performance improves by \textbf{+0.5\%} over MoCo-v2 at 800 epochs. SynCo also transfers well to a range of downstream tasks.
\end{abstract}

\section{Introduction}
\label{sec:introduction}

\begin{figure}
\centering
\includegraphics[width=0.4\textwidth]{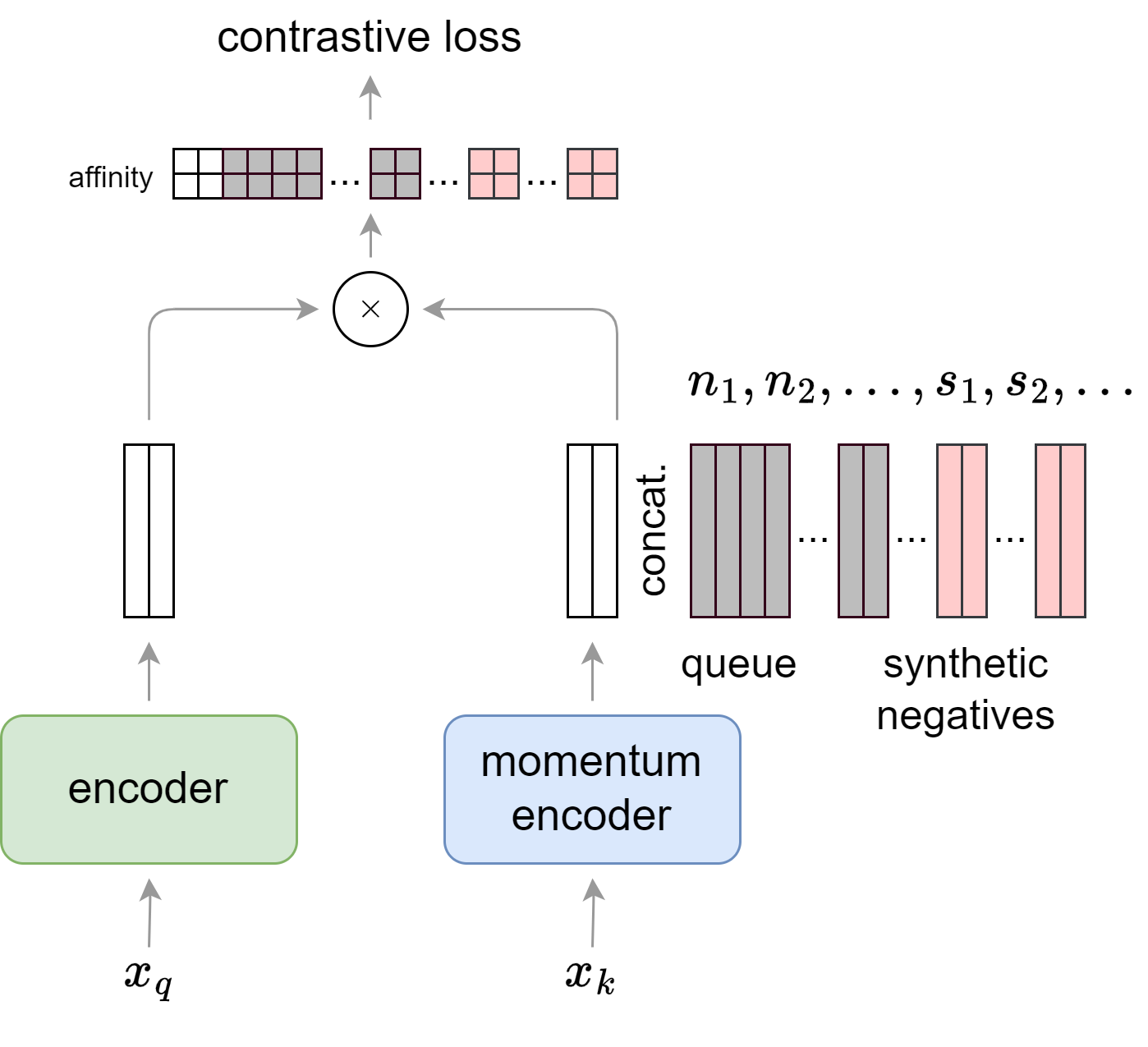}
\caption{\textbf{SynCo extends MoCo}~\cite{he2020moco, chen2020mocov2} by introducing synthetic hard negatives generated ``on-the-fly'' from a memory queue. Two augmented views $\mathbf{x}_q$ and $\mathbf{x}_k$ are processed by an encoder and momentum encoder to produce features $\mathbf{q}$ and $\mathbf{k}$. Synthetic hard negatives $\mathbf{s}_1, \mathbf{s}_2, \ldots$ generated using SynCo strategies are concatenated with queue negatives $\mathbf{n}_1, \mathbf{n}_2, \ldots$ to compute the affinity matrix for InfoNCE loss with the positive pair ($\mathbf{q}$, $\mathbf{k}$).}
\label{fig:framework}
\end{figure}

Contrastive self-supervised learning has become a dominant paradigm for visual representation learning, achieving strong performance across many downstream tasks without requiring labeled data \cite{chen2020simclr, he2020moco, grill2020byol, caron2021dino}. The core mechanism is instance discrimination \cite{wu2018instdis}: augmented views of an image form a positive pair, while views from other images are negatives, and the encoder learns to align positives while repelling negatives via the InfoNCE objective \cite{oord2019cpc}. The quality of learned representations depends critically on the difficulty of the negative samples: easy negatives saturate early and contribute negligible gradient signal, whereas hard negatives (semantically close to the query) force the encoder to learn fine-grained discriminative features \cite{robinson2021contrastivelearninghardnegative, kalantidis2020mochi}.

Two broad strategies enrich the pool of hard negatives. The first increases input-space diversity through stronger or more varied augmentations \cite{caron2020swav, tian2020infomin, zhang2018mixup, balestriero2023cookbook}, but generating additional views requires extra encoder forward passes, making it considerably more expensive than a standard iteration. The second synthesizes hard negatives directly in the representation space, operating on cached queue embeddings with no additional forward passes, so the gains come practically for free. To our knowledge only MoCHi \cite{kalantidis2020mochi} has pursued this direction, via convex combinations of existing negatives. However, its mixing is restricted to the geometric family and does not consistently improve over MoCo-v2 (\Cref{tab:lineval_imagenet_200ep,tab:lineval_imagenet_800ep}), motivating a more comprehensive approach to representation-space augmentation.

In this paper we propose SynCo (\underline{\textbf{{Sy}}}nthetic \underline{\textbf{{n}}}egatives in \underline{\textbf{{Co}}}ntrastive learning), which synthesizes hard negatives directly in the representation space at negligible cost. Rather than mining hard negatives from data, SynCo selects the most challenging negatives from the queue by cosine similarity to the query and synthesizes new, harder ones ``on-the-fly'' using six complementary strategies, each targeting a distinct region of the representation space.

The main \textbf{contributions} are as follows: \textbf{(i)} SynCo generates synthetic hard negatives with \textit{six} complementary strategies spanning the canonical perturbation families (\cf \Cref{subsec:generation}): geometric, stochastic, and adversarial, with no additional encoder forward passes. \textbf{(ii)} These strategies extend MoCHi \cite{kalantidis2020mochi} beyond the convex hull of existing negatives via gradient-guided perturbations and controlled noise, improving uniformity (\Cref{fig:align_uniform}) and inter-/intra-class distances (\Cref{fig:class_concentration}). \textbf{(iii)} Unlike MoCHi, SynCo scales with pretraining duration: with a simple synthetic negative schedule it gains \textbf{+0.5\%} over MoCo-v2 at 800 epochs (\Cref{tab:lineval_imagenet_800ep}), where MoCHi trails MoCo-v2 by 2.4\%. \textbf{(iv)} It improves consistently across downstream tasks: linear evaluation (\Cref{tab:lineval_imagenet_200ep,tab:lineval_imagenet_800ep}), semi-supervised learning (\Cref{tab:semisup_imagenet}), and object detection (\Cref{tab:detection_coco_voc}).

\section{Related Work}
\label{sec:relatedwork}

We review contrastive self-supervised learning, data augmentation, and hard negative mining.

\subsection{Self-Supervised Contrastive Learning}
\label{subsec:contrastive_learning_related}

Contrastive learning treats each image as its own class for instance discrimination \cite{chen2020simclr, he2020moco}, pulling an anchor and positive together while pushing negatives apart \cite{khosla2021supcon}. Positive pairs come from multiple views \cite{tian2020infomin, caron2020swav}, using color decomposition \cite{tian2020cmc}, random augmentation \cite{chen2020simclr, he2020moco}, image patches \cite{oord2019cpc}, or student-teacher representations \cite{grill2020byol, caron2021dino, oquab2023dinov2}. The common objective, InfoNCE \cite{oord2019cpc} or its variants \cite{chen2020simclr, dwibedi2021nnclr, tomasev2022relicv2, yeh2022dcl}, maximizes mutual information \cite{hjelm2019dim, bachman2019amdim}. SimCLR uses large batches \cite{chen2020simclr}, whereas MoCo \cite{he2020moco, chen2020mocov2}, PIRL \cite{misra2019pirl}, and InstDis \cite{wu2018instdis} employ memory structures. Other lines add regularizers \cite{mitrovic2020relic, bardes2022vicreg, zhu2022tico, bardes2022vicregl}, prevent collapse via redundancy reduction \cite{zbontar2021barlowtwins, bandara2023mixedbarlowtwins}, remove negatives through asymmetric Siamese structures \cite{grill2020byol, chen2020simsiam, caron2021dino, oquab2023dinov2}, address false-negative pairs \cite{zhuang2019la, li2021pcl}, or decouple feature and metric learning \cite{yeh2022dcl}. SynCo targets the InfoNCE family specifically and is orthogonal to negative-free methods, which operate in a fundamentally different paradigm.

\subsection{Data Augmentation}
\label{subsec:input_augmentation_related}

Data augmentation is central to contrastive learning, since the transformation choice determines the learned invariances \cite{chen2020simclr, balestriero2023cookbook}. Beyond standard augmentations, work has sought richer input-space diversity: multi-crop \cite{caron2020swav} adds smaller-scale views, mutual-information view selection \cite{tian2020infomin} and nearest-neighbour positive mining \cite{dwibedi2021nnclr} improve pair semantics, learned policies such as RandAugment \cite{cubuk2019randaugment} search transformation sequences, and generative approaches synthesize entirely new images \cite{jahanian2022generative}. Pixel-level mixing such as Mixup \cite{zhang2018mixup} and CutMix \cite{yun2019cutmix} interpolates between images, while CLSA \cite{wang2022clsa} and SaSSL \cite{rojasgomez2024sassl} further explore augmentation-driven SSL. All input-space methods share one cost: generating diverse views requires extra encoder forward passes or auxiliary networks, making them far more expensive than operating on embeddings.

\subsection{Hard Negatives}
\label{subsec:hard_negatives_related}

Hard negative mining selects samples close to the anchor in representation space but belonging to different instances. MoCo \cite{he2020moco} maintains challenging negatives via dynamic queues and momentum updates, while SimCLR \cite{chen2020simclr} and InfoMin \cite{tian2020infomin} adjust difficulty through augmentation. Recent work examines optimal sampling strategies \cite{robinson2021contrastivelearninghardnegative, jiang2025hardnegative}, semantic negative sampling \cite{ge2022robustcontrastivelearningusing}, and the broader impact of negative sampling choices \cite{yang2024doesnegativesamplingmatter}. Closest to us, MoCHi \cite{kalantidis2020mochi} synthesizes harder negatives directly in representation space via convex combinations of existing negative features, avoiding additional forward passes. While computationally appealing, it covers only the geometric family, constraining synthesis to the convex hull of existing negatives and leaving the stochastic and adversarial families unexplored. SynCo extends this direction to all three canonical perturbation families, geometric, stochastic, and adversarial, covering a substantially larger and more diverse region of representation space while preserving the negligible computational overhead.

\section{Preliminaries}
\label{sec:preliminaries}

We establish the foundations of contrastive learning and analyze the critical role of hard negatives in representation learning.

\subsection{Contrastive Learning}
\label{subsec:contrastive_learning}

Contrastive learning aligns similar pairs and separates dissimilar ones, often framed as a dictionary look-up trained with a contrastive loss \cite{he2020moco}. Given an image $x$ and an augmentation distribution $\mathcal{T}$, we create two views via transformations $t_q, t_k \sim \mathcal{T}$, \ie $x_q = t_q(x)$ and $x_k = t_k(x)$. Two encoders, the query $f_q$ and key $f_k$, generate vectors $\mathbf{q}=f_q(x_q)$ and $\mathbf{k}=f_k(x_k)$. The objective minimizes a contrastive loss using the InfoNCE criterion \cite{oord2019cpc}:

\begin{equation}\label{eq:loss_contrastive}
\mathcal{L}(\mathbf{q},\mathbf{k},\mathcal{Q}) = -\log \frac{\exp(\mathbf{q}^\top \cdot \mathbf{k} / \tau )}{\exp(\mathbf{q}^\top \cdot \mathbf{k} / \tau ) + \sum\limits_{\mathbf{n} \in \mathcal{Q}} \exp(\mathbf{q}^\top \cdot \mathbf{n} / \tau)}
\end{equation}

\noindent Here, $\mathbf{k}$ is $f_k$'s output from the same image as $\mathbf{q}$, and $\mathcal{Q} = \{\mathbf{n}_i\}_{i=1}^{K}$ contains $K$ negatives from different images. The temperature $\tau$ scales the $\ell_2$-normalized vectors $\mathbf{q}$ and $\mathbf{k}$. The key encoder $f_k$ is updated either synchronously with $f_q$, sharing identical weights \cite{chen2020simclr}, or via a momentum update $\theta_k \leftarrow m \cdot \theta_k + (1-m) \cdot \theta_q$ \cite{he2020moco}, where $m \in [0, 1]$ is the momentum coefficient; the latter lets $f_k$ evolve slowly, providing consistent negatives and stabilizing learning. The memory bank $\mathcal{Q}$ can be an external memory of all dataset images \cite{misra2019pirl, tian2020cmc, wu2018instdis}, a queue of recent batches \cite{he2020moco}, or the current minibatch \cite{chen2020simclr}. The gradient of the loss in \Cref{eq:loss_contrastive} with respect to $\mathbf{q}$ is:

\begin{equation}\label{eq:loglikehood}
\frac{\partial \mathcal{L}(\mathbf{q},\mathbf{k},\mathcal{Q})}{\partial \mathbf{q}} = -\frac{1}{\tau} \left( (1 - p_k) \cdot \mathbf{k} - \sum_{\mathbf{n} \in \mathcal{Q}} p_n \cdot \mathbf{n} \right)
\end{equation}

where

\begin{equation}
p_{z_i} = \frac{\exp(\mathbf{q}^\top \cdot \mathbf{z_i} / \tau)}{\sum_{j \in Z} \exp(\mathbf{q}^\top \cdot \mathbf{z_j} / \tau)}
\end{equation}

\noindent with $\mathbf{z_i} \in \mathcal{Q} \cup \{\mathbf{k}\}$. The positive and negative logits contribute to the loss similarly to a $(K + 1)$-way cross-entropy classification, with the key logit representing the query's latent class \cite{arora2019theoretical}.

\subsection{Understanding Hard Negatives}
\label{subsec:understanding_hard_neg}

\begin{figure*}[!t]
    \centering
    \begin{subfigure}[t]{0.32\textwidth}
        \centering
        \includegraphics[width=\linewidth]{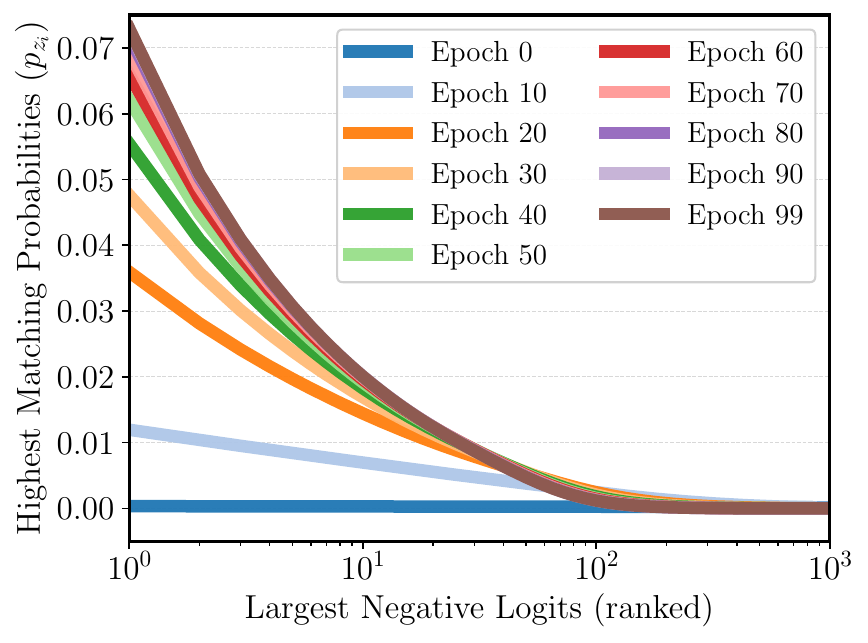}
        \caption{}
        \label{fig:histogram_probs}
    \end{subfigure}
    \hfill
    \begin{subfigure}[t]{0.32\textwidth}
        \centering
        \includegraphics[width=\linewidth]{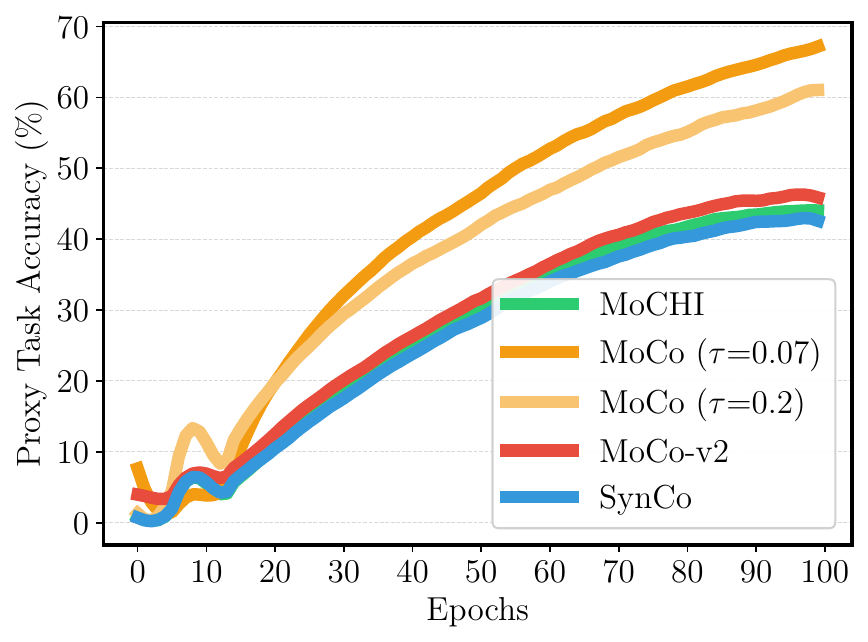}
        \caption{}
        \label{fig:proxy_task_acc_moco}
    \end{subfigure}
    \hfill
    \begin{subfigure}[t]{0.32\textwidth}
        \centering
        \includegraphics[width=\linewidth]{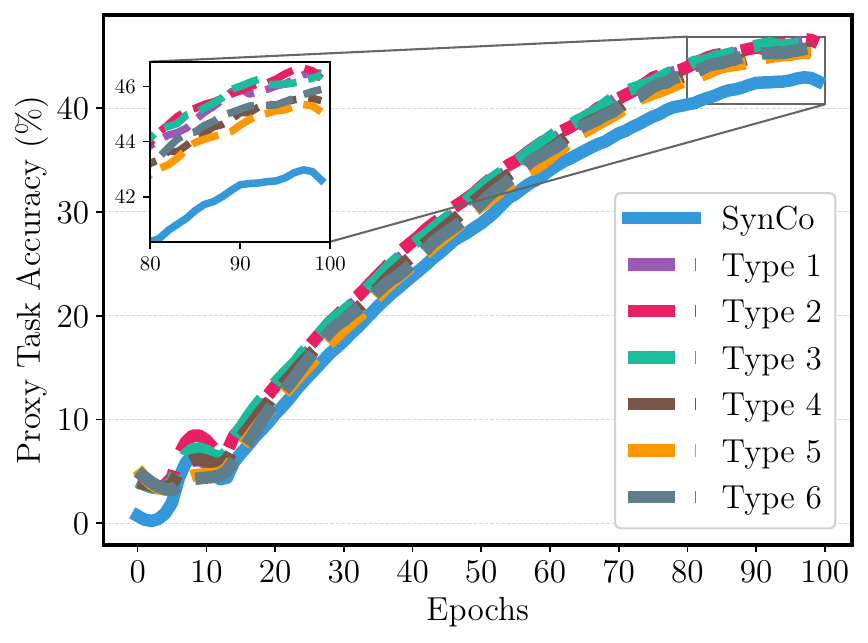}
        \caption{}
        \label{fig:proxy_task_acc_synco}
    \end{subfigure}
    \caption{\textbf{ImageNet-100 experiments.}
    \subref{fig:histogram_probs} Histogram of the top 1024 matching probabilities $p_{z_i}$, $z_i \in \mathcal{Q}$ for MoCo-v2 over various epochs. Logits are organized in descending order, and each line indicates the mean matching probability across all queries.
    \subref{fig:proxy_task_acc_moco} Performance comparison of MoCo, MoCo-v2, MoCHi, SynCo in terms of accuracy on the proxy task (percentage of queries where the key is ranked higher than all negatives).
    \subref{fig:proxy_task_acc_synco} Performance comparison of SynCo under various configurations in terms of accuracy on the proxy task.}
    \label{fig:contrastive_analysis}
\end{figure*}

The effectiveness of contrastive learning hinges critically on the use of hard negatives \cite{arora2019theoretical, hadsell2006dimensionality, iscen2018mining, mishchuk2017working, wu2018instdis, kalantidis2020mochi}, yet current approaches struggle to leverage them efficiently. Sampling within a batch requires larger batch sizes \cite{chen2020simclr, chen2021mocov3}, while maintaining a full-dataset memory bank incurs substantial overhead in keeping it up-to-date \cite{misra2019pirl, wu2018instdis, he2020moco, chen2020mocov2}. These limitations motivate more efficient strategies for generating and using hard negatives.

\paragraph{\bf Hardness of negatives.} The ``\textit{hardness}'' of a negative, defined by its similarity to positive samples in the representation space, determines how challenging it is to differentiate. \Cref{fig:histogram_probs} illustrates the evolution of negative hardness during MoCo-v2 training: the probability distribution is initially near-uniform, but as training progresses fewer negatives contribute significantly to the loss, indicating the model rapidly learns to distinguish most of them and leaves only a small challenging subset. SynCo addresses this by synthesizing new hard negatives on-the-fly, replenishing the pool of informative contrasts throughout training \cite{kalantidis2020mochi}.

\paragraph{\bf Difficulty of the proxy task.} The ``\textit{difficulty}'' of the proxy task, set by the self-supervised objective, significantly influences representation quality. \Cref{fig:proxy_task_acc_moco} compares the proxy task performance of MoCo and MoCo-v2 on ImageNet-100, measured by the percentage of queries where the key ranks above all negatives. MoCo-v2, with more aggressive augmentations, shows lower proxy task performance than MoCo, indicating a harder objective, yet paradoxically this difficulty correlates with better downstream performance such as linear classification \cite{kalantidis2020mochi}. \Cref{fig:proxy_task_acc_synco} further shows how SynCo varies across configurations, with the complete framework consistently outperforming each individual strategy ($S^1$ to $S^6$) in isolation. This counterintuitive link suggests that harder self-supervised objectives yield more robust and transferable representations, motivating strategies that dynamically modulate task difficulty during training.

\section{Synthetic Hard Negatives for Contrastive Learning}
\label{sec:synco}

We describe SynCo's six synthetic negative strategies, all operating on cached queue embeddings with no additional encoder forward passes. They span the canonical perturbation families: geometric ($S^1, S^2, S^3$), stochastic ($S^4$), and adversarial ($S^5, S^6$). Building on MoCHi \cite{kalantidis2020mochi}, we add four strategies that explore complementary regions beyond the convex hull of existing negatives.

\subsection{Generating Synthetic Hard Negatives}
\label{subsec:generation}

Let $\mathbf{q}$ represent the query image, $\mathbf{k}$ its corresponding key, and $\mathbf{n} \in \mathcal{Q}$ denote the negative features from a memory structure of size $K$. The loss associated with the query is computed using the logits $\ell (\mathbf{z_i}) = \mathbf{q}^\top \cdot \mathbf{z}_i / \tau$, processed through a softmax. We define $\mathcal{\hat{Q}}=\{\mathbf{n}_1, \mathbf{n}_2, \ldots, \mathbf{n}_K\}$ as the ordered set of all negative features, where $\ell(\mathbf{n}_i) > \ell (\mathbf{n}_j)$ for all $i < j$, so negatives are sorted by decreasing similarity to the query. The most challenging negatives are selected by truncating $\mathcal{\hat{Q}}$ to its first $N < K$ elements, denoted $\mathcal{\hat{Q}}^N$. Let $S^{(i)} = \{\mathbf{s}_1^{(i)}, \mathbf{s}_2^{(i)}, \ldots, \mathbf{s}_{N_i}^{(i)}\}$ denote the set of synthetic negatives generated for type $i$, where $i \in \{1, 2, 3, 4, 5, 6\}$ indexes the six strategies and $N_i$ is the number of synthetic negatives generated for the $i$-th strategy. All synthetic hard negatives are $\ell_2$-normalised before being added to the set of negative logits for the query.

\paragraph{\bf Interpolated synthetic negatives ($S^1$).}
Our first strategy creates synthetic negatives via controlled interpolation between samples, similar to MoCHi's type 2 \cite{kalantidis2020mochi}, generating features that lie between the query and existing hard negatives. For each query $\mathbf{q}$, we generate $N_1$ synthetic hard negatives by mixing $\mathbf{q}$ with a randomly chosen feature from the $N$ hardest negatives in $\mathcal{\hat{Q}}^N$. A synthetic negative $\mathbf{s}_k^1 \in S^1$ is given by:

\begin{equation}
\mathbf{s}_k^1 = \alpha_k \cdot \mathbf{q} + (1 - \alpha_k) \cdot \mathbf{n}_i, \quad \alpha_k \in (0, \alpha_{\text{max}})
\end{equation}

\noindent where $\mathbf{n}_i \in \mathcal{\hat{Q}}^N$ and $\alpha_k \sim \mathcal{U}(0, \alpha_{\text{max}})$. We set $\alpha_{\text{max}} = 0.5$ so the query's contribution is always smaller than the negative's.

\paragraph{\bf Extrapolated synthetic negatives ($S^2$).}
As a natural extension, we extrapolate beyond the query in the direction of the hardest negative, controlling the magnitude through a coefficient to keep the synthetic negatives challenging but not intractable. For each query $\mathbf{q}$, we generate $N_2$ hard negatives using a randomly chosen feature from the $N$ hardest negatives in $\mathcal{\hat{Q}}^N$. Then $\mathbf{s}_k^2 \in S^2$ is given by:

\begin{equation}
\mathbf{s}_k^2 = \mathbf{n}_i + \beta_k \cdot (\mathbf{n}_i - \mathbf{q}), \quad \beta_k \in (1, \beta_{\text{max}})
\end{equation}

\noindent where $\mathbf{n}_i \in \mathcal{\hat{Q}}^N$ and $\beta_k \sim \mathcal{U}(1, \beta_{\text{max}})$. We choose $\beta_{\text{max}} = 1.5$.

\paragraph{\bf Mixup synthetic negatives ($S^3$).}
We generate challenging negatives by combining pairs of hard negatives, similar to MoCHi's type 1 \cite{kalantidis2020mochi}. For each query $\mathbf{q}$, we generate $N_3$ hard negatives by combining pairs of the $N$ hardest negatives in $\mathcal{\hat{Q}}^N$. For $\mathbf{s}_k^3 \in S^3$ we have:

\begin{equation}
\mathbf{s}_k^3 = \gamma_k \cdot \mathbf{n}_i + (1 - \gamma_k) \cdot \mathbf{n}_j, \quad \gamma_k \in (0, 1)
\end{equation}

\noindent where $\mathbf{n}_i, \mathbf{n}_j \in \mathcal{\hat{Q}}^N$ and $\gamma_k \sim \mathcal{U}(0, 1)$.

\paragraph{\bf Noise-injected synthetic negatives ($S^4$).}
To prevent overfitting to specific negative patterns while preserving hard-negative characteristics, we introduce controlled stochasticity via noise injection. For each query $\mathbf{q}$, we generate $N_4$ hard negatives by adding Gaussian noise to the hardest negatives. Each $\mathbf{s}_k^4 \in S^4$ is given by:

\begin{equation}
\mathbf{s}_k^4 = \mathbf{n}_i + \mathcal{N}(\mathbf{0}, \sigma^2 \cdot \mathbf{I})
\end{equation}

\noindent where $\mathbf{n}_i \in \mathcal{\hat{Q}}^N$ and $\mathcal{N}(\mathbf{0}, \sigma^2 \cdot \mathbf{I})$ is Gaussian noise with standard deviation $\sigma$ ($\mathbf{I}$ is the identity matrix).

\paragraph{\bf Perturbed synthetic negatives ($S^5$).}
Drawing on adversarial training \cite{mehrabi21a}, we introduce perturbed negatives using gradient-based perturbations with variable magnitude. For each query $\mathbf{q}$, we generate $N_5$ hard negatives by perturbing the hardest negatives. Each $\mathbf{s}_k^5 \in S^5$ is given by:

\begin{equation}
\mathbf{s}_k^5 = \mathbf{n}_i + \delta \cdot \nabla_{\mathbf{n}_i} \text{sim}(\mathbf{q}, \mathbf{n}_i)
\end{equation}

\noindent where $\mathbf{n}_i \in \mathcal{\hat{Q}}^N$, $\text{sim}(\cdot, \cdot)$ is the similarity function, and $\delta$ controls the perturbation magnitude. Unlike fixed interpolation, this generalizes to arbitrary similarity functions and yields negatives of varying hardness.

\paragraph{\bf Adversarial synthetic negatives ($S^6$).}
While similar to $S^5$, adversarial negatives differ in their gradient scaling. For each query $\mathbf{q}$, we generate $N_6$ hard negatives by applying adversarial perturbations to the hardest negatives to maximize their similarity to the query. Each $\mathbf{s}_k^6 \in S^6$ is defined as:

\begin{equation}
\mathbf{s}_k^6 = \mathbf{n}_i + \eta \cdot \text{sign}(\nabla_{\mathbf{n}_i} \text{sim}(\mathbf{q}, \mathbf{n}_i))
\end{equation}

\noindent where $\mathbf{n}_i \in \mathcal{\hat{Q}}^N$ and $\eta$ controls the perturbation magnitude. Whereas $S^5$ allows variable perturbation sizes, $S^6$ enforces unit magnitude through the sign function, producing the most challenging contrasts.

\subsection{Integrating Synthetic Hard Negatives into the Contrastive Loss}
\label{subsec:integration}

The synthetic hard negatives are integrated by modifying the InfoNCE loss. Let $\mathcal{S} = \bigcup_{i=1}^{6} S^{(i)}$ be the concatenation of all synthetic hard negatives, where $S^{(i)}$ is the set generated by the $i$-th strategy. This set augments the original negatives $\mathcal{Q}$, providing more diverse and challenging contrasts for the query. The modified InfoNCE loss is:

\begin{equation}
    \mathcal{L}(\mathbf{q}, \mathbf{k}, \mathcal{Q}, \mathcal{S}) = -\log \frac{\exp(\mathbf{q}^\top \cdot \mathbf{k} / \tau)}{\exp(\mathbf{q}^\top \cdot \mathbf{k} / \tau) + Z}
\end{equation}

\noindent where $Z$ represents the negative samples:

\begin{equation}
Z = \tikz[baseline]{\node[anchor=base] (n) {$\sum\limits_{\substack{\mathbf{n} \in \mathcal{Q}}} \exp(\mathbf{q}^\top \cdot \mathbf{n} / \tau)$};\draw[decorate, decoration={brace, mirror, amplitude=5pt}, thick, black] (n.south west) -- (n.south east) node[midway, below=6pt, black] {memory-based negatives};} + \tikz[baseline]{\node[anchor=base] (s) {$\sum\limits_{\substack{\mathbf{s} \in \mathcal{S}}} \exp(\mathbf{q}^\top \cdot \mathbf{s} / \tau)$};\draw[decorate, decoration={brace, mirror, amplitude=5pt}, thick, cvprblue] (s.south west) -- (s.south east) node[midway, below=6pt, cvprblue] {synthetic negatives};}.
\end{equation}

\noindent Here, $\mathcal{Q}$ is the original memory-based negatives and $\mathcal{S}$ the synthetic ones; together they expose the model to a wider variety of challenging examples, encouraging more robust and generalizable representations. The overhead is negligible: the six strategies add only $\sum_{i=1}^{6} N_i \ll K$ embeddings, computed on-the-fly in parallel with the forward pass, and are never stored, so the memory footprint is unchanged. %\Cref{appendix:implementation_details} provides a detailed description of the SynCo algorithm (\cf \Cref{alg:synco,alg:type1,alg:type2,alg:type3,alg:type4,alg:type5,alg:type6}).

\section{Experiments}
\label{sec:experiments}

In this section, we demonstrate SynCo's effectiveness on ImageNet linear evaluation, semi-supervised learning, and transfer learning to object detection. %The Appendix contains: \textbf{(i}) implementation details (\Cref{appendix:implementation_details}); \textbf{(ii)} extended ablations ($\sigma$, $\delta$, $\eta$, $N$, $N_i$, $K$) (\Cref{appendix:ablations}); \textbf{(iv)} vision transformers \cite{dosovitskiy2021vit,touvron2021deit} (\Cref{appendix:synco_on_vits}); \textbf{(v)} robustness (\Cref{appendix:robustness_ood,appendix:robustness_adversarial}); \textbf{(vi)} visualizations, \eg, t-SNE (\Cref{appendix:tsne}), GradCAM (\Cref{appendix:gradcam}), UMAP (\Cref{appendix:umap}), nearest neighbor retrieval (\Cref{appendix:nn}).

\subsection{Implementation Details}
\label{subsec:implementation_details}

We pretrain SynCo on ImageNet ILSVRC-2012 \cite{deng2009imagenet} using a ResNet-50 encoder \cite{he2015resnet}. Our method builds upon MoCo-v2 \cite{chen2020mocov2}; thus, it is only \textit{fair} to compare against other MoCo-based methods \cite{chen2020mocov2,li2021pcl,kalantidis2020mochi,yeh2022dcl}, which share similar architectures and training setups (\cf \textbf{bold} entries in \Cref{tab:lineval_imagenet_200ep,tab:lineval_imagenet_800ep,tab:semisup_imagenet,tab:detection_coco_voc}, indicating the best performance among MoCo-based methods; we \textit{underline} the second best). For training we use $K = 65$k. For SynCo, we also have a warm-up of 10 epochs, \ie for the first epochs we do \textit{not} synthesize hard negatives. We empirically set SynCo's hyperparameters $\sigma$, $\delta$, $\eta$ to 0.01. For hard negative generation, we use the top $N=1024$ hardest negatives, with $N_1=N_2=N_3=256$ and $N_4=N_5=N_6=64$. For ImageNet linear evaluation, we train a linear classifier on frozen features for 100 epochs, using a batch size of 256 and a cosine learning rate schedule ($lr=30.0$). To evaluate transfer learning, we apply SynCo to object detection tasks. For PASCAL VOC \cite{everingham2009voc}, we fine-tune a Faster R-CNN \cite{ren2016fasterrcnn} on \texttt{trainval07+12} and test on \texttt{test2007}. For COCO \cite{li2015coco}, we use a Mask R-CNN \cite{he2018maskrcnn}, fine-tuning on \texttt{train2017} and evaluating on \texttt{val2017}. We employ Detectron2 \cite{wu2019detectron2} and report standard AP metrics, following \cite{he2020moco} \textit{without} additional tuning.

\subsection{Linear Evaluation on ImageNet}
\label{subsec:imagenet_eval}

\begin{table}[!t]
\centering
\caption{\textbf{Linear evaluation on ImageNet ILSVRC-2012.} Top-1 accuracy (in \%) with 200 epochs of pretraining using ResNet-50. Results for SynCo are given as avg./max over 3 runs.}
\label{tab:lineval_imagenet_200ep}
\begin{tabular}{lc}
\toprule
\textbf{Method} & \textbf{Top-1} \\
\midrule
\textit{Supervised} & 76.5 \\
PIRL~\citep{misra2019pirl} & 63.6 \\
LA~\citep{zhuang2019la} & 60.2 \\
CMC~\citep{tian2020cmc} & 60.0 \\
% InfoMin~\citep{tian2020infomin} & 70.1 \\
SimSiam~\citep{chen2020simsiam} & 68.1 \\
ReSSL~\citep{zheng2021ressl} & 62.9 \\
AdCo~\citep{hu2021adco} & 68.6 \\
SimCLR + DCL~\citep{yeh2022dcl} & 65.8 \\
\midrule
\multicolumn{2}{c}{\textit{MoCo-based}} \\
MoCo~\citep{he2020moco} & 60.7 \textcolor{red}{\small{$\downarrow$6.8}} \\
PCL-v1~\citep{li2021pcl} & 61.5 \textcolor{red}{\small{$\downarrow$6.0}} \\
MoCo-v2~\citep{chen2020mocov2} (\textit{baseline}) & 67.5 \textcolor{gray}{\small{$\uparrow$0.0}} \\
MoCHI~\citep{kalantidis2020mochi} & 66.9 \textcolor{red}{\small{$\downarrow$0.6}} \\
PCL-v2~\citep{li2021pcl} & 67.6 \textcolor{green}{\small{$\uparrow$0.1}} \\
MoCo-v2 + DCL~\citep{yeh2022dcl} & 67.6 \textcolor{green}{\small{$\uparrow$0.1}} \\
MoCo-v2 + NS~\citep{ge2022robustcontrastivelearningusing} & \underline{67.9} \textcolor{green}{\small{$\uparrow$0.4}} \\
SynCo (ours) & \cellcolor{pearDarker!08} \textbf{67.9}/\textbf{68.1} \textcolor{green}{\small{$\uparrow$0.6}} \\
\bottomrule
\end{tabular}
\end{table}

\begin{table}[!t]
\centering
\caption{\textbf{Linear evaluation on ImageNet ILSVRC-2012.} Top-1 and top-5 accuracies (in \%) for models trained with \textit{extended epochs} using ResNet-50. Results for SynCo are based on 1 run. \textbf{Symbols}: $\dag$ We stop generating synthetic negatives at epoch 400.}
\label{tab:lineval_imagenet_800ep}
\begin{tabular}{lccc}
\toprule
\textbf{Method} & \textbf{Epochs} & \textbf{Top-1} & \textbf{Top-5} \\
\midrule
InfoMin~\citep{tian2020infomin} & 800 & 73.0 & 91.1 \\
SimSiam~\citep{chen2020simsiam} & 800 & 68.1 & - \\
SimCLR~\citep{chen2020simclr} & 1000 & 69.3 & - \\
BYOL~\citep{grill2020byol} & 1000 & 74.3 & 91.6 \\
DINO~\citep{caron2021dino} & 800 & 75.3 & - \\
Barlow Twins~\citep{zbontar2021barlowtwins} & 1000 & 73.2 & 91.0 \\
AdCo~\citep{hu2021adco} & 800 & 72.8 & - \\
VICReg~\citep{bardes2022vicreg} & 1000 & 73.2 & 91.1 \\
CaCo~\citep{wang2022caco} & 800 & 74.1 & - \\
All4One~\citep{estepa2023all4one} & 800 & 66.6 & 87.5 \\
\midrule
\multicolumn{4}{c}{\textit{MoCo-based}} \\
MoCo-v2~\citep{chen2020mocov2} & 800 & \underline{71.1} \textcolor{gray}{\small{$\uparrow$0.0}} & 90.1 \\
MoCHI~\citep{kalantidis2020mochi} & 800 & 68.7 \textcolor{red}{\small{$\downarrow$2.4}} & - \\
SynCo (ours) & 800 & 70.7 \cellcolor{pearDarker!08} \textcolor{red}{\small{$\downarrow$0.4}} & \cellcolor{pearDarker!08} 89.8 \\
SynCo$^\dag$ (ours) & 800 & \cellcolor{pearDarker!08} \textbf{71.6} \textcolor{green}{\small{$\uparrow$0.5}} & \cellcolor{pearDarker!08} 90.5 \\
\bottomrule
\end{tabular}
\end{table}
\begin{table}[!t]
\centering
\setlength{\tabcolsep}{1.5mm}
\caption{\textbf{Semi-supervised learning on ImageNet ILSVRC-2012.} Top-1 and top-5 accuracies with 1\% and 10\% training examples using ResNet-50. Results for SynCo are averaged over 3 runs.}
\label{tab:semisup_imagenet}
\begin{tabular}{lccccc}
\toprule
\multirow{2}{*}{\textbf{Method}} & \multirow{2}{*}{\textbf{Epochs}} & \multicolumn{2}{c}{\textbf{Top-1}} & \multicolumn{2}{c}{\textbf{Top-5}} \\
\cmidrule(lr){3-4} \cmidrule(lr){5-6}
& & 1\% & 10\% & 1\% & 10\% \\
\midrule
\textit{Supervised} & & 25.4 & 56.4 & 48.4 & 80.4 \\
InstDis \citep{wu2018instdis} & 200 & - & - & 39.2 & 77.4 \\
PIRL \citep{misra2019pirl} & 800 & 30.7 & 60.4 & 57.2 & 83.8 \\
SimCLR \citep{chen2020simclr} & 1000 & 48.3 & 65.6 & 75.5 & 87.8 \\
BYOL \citep{grill2020byol} & 1000 & 53.2 & 68.8 & 78.4 & 89.0 \\
SwAV \citep{caron2020swav} & 800 & 53.9 & 70.2 & 78.5 & 89.9 \\
Barlow Twins \citep{zbontar2021barlowtwins} & 1000 & 55.0 & 69.7 & 79.2 & 89.3 \\
%PAWS \citep{assran2021paws} & 200 & 63.8 & 73.9 & - & - \\
%MSF \citep{koohpayegani2021msf} & 800 & 55.5 & 79.9 & 66.5 & 87.6 \\
VICReg \citep{bardes2022vicreg} & 1000 & 54.8 & 69.5 & 79.4 & 89.5 \\
All4One \citep{estepa2023all4one} & 800 & 39.0 & - & 60.0 & - \\
\midrule
\multicolumn{6}{c}{\textit{MoCo-based}} \\
MoCo-v2 (repr.) & 800 & 48.2 & 66.1 & 75.8 & 87.6 \\
PCL-v1 \citep{li2021pcl} & 200 & - & - & 75.3 & 85.6 \\
PCL-v2 \citep{li2021pcl} & 200 & - & - & 73.9 & 85.0 \\
MoCHi (repr.) & 800 & 50.4 & 65.7 & 76.2 & 87.2 \\
SynCo (ours) & 800 & \cellcolor{pearDarker!08}\underline{50.8} & \cellcolor{pearDarker!08}\underline{66.6} & \cellcolor{pearDarker!08}\underline{77.5} & \cellcolor{pearDarker!08}\underline{88.0} \\
SynCo$^\dag$ (ours) & 800 & \cellcolor{pearDarker!08}\textbf{51.2} & \cellcolor{pearDarker!08}\textbf{67.1} & \cellcolor{pearDarker!08}\textbf{78.0} & \cellcolor{pearDarker!08}\textbf{88.7} \\
\bottomrule
\end{tabular}
\end{table}

We train a linear classifier on frozen ImageNet-pretrained features. At 200 epochs, SynCo obtains 67.9\% $\pm$ 0.16\% top-1 and 88.0\% $\pm$ 0.05\% top-5, with clear gains over MoCo-based methods (\textbf{+0.4\%} over MoCo-v2, \textbf{+1.0\%} over MoCHi, \textbf{+0.3\%} over PCL-v2 and DCL); unlike MoCHi, which trails MoCo-v2, our synthetic negatives give consistent and stable improvements. At 800 epochs, SynCo reaches 70.7\% top-1 (\textbf{+2.0\%} over MoCHi) but, like MoCHi, does not surpass MoCo-v2, likely due to an overly hard proxy task. Since performance plateaus around epoch 400, we \textit{stop} synthesizing negatives thereafter. This two-phase schedule follows the easy-to-hard curriculum principle \cite{lin2024accelerating}: synthetic negatives tighten decision boundaries early, while the cooldown re-anchors the model on the natural data manifold before convergence, yielding 71.6\% top-1, a \textbf{+0.5\%} improvement over MoCo-v2.

In a semi-supervised setting with 1\% and 10\% of labeled ImageNet data (\Cref{tab:semisup_imagenet}), SynCo reaches 50.8\% $\pm$ 0.21\% top-1 (\textbf{+25.4\%} over supervised, \textbf{+2.6\%} over MoCo-v2, \textbf{+2.5\%} over SimCLR) and 77.5\% $\pm$ 0.12\% top-5 with 1\% labels, and 66.6\% $\pm$ 0.19\% top-1 (\textbf{+10.2\%} over supervised, \textbf{+0.5\%} over MoCo-v2, \textbf{+1.0\%} over SimCLR) and 88.0\% $\pm$ 0.10\% top-5 with 10\% labels. Applying the epoch-400 cooldown improves these further to 51.2\% $\pm$ 0.23\% / 78.0\% $\pm$ 0.14\% (1\% labels) and 67.1\% $\pm$ 0.20\% / 88.7\% $\pm$ 0.11\% (10\% labels).

\subsection{Transferring to Detection}
\label{subsec:detection}

\begin{table*}[!t]
\centering
\caption{\textbf{Transfer learning on COCO using R50-C4 with $1\times$ and $2\times$ training schedules and PASCAL VOC07+12 using R50-C4.} We report standard AP metrics, \ie, $\text{AP}^{\text{bb}}$ (bounding box detection) and $\text{AP}^{\text{m}}$ (instance segmentation). Results for SynCo are averaged over 3 runs.}
\label{tab:detection_coco_voc}
\setlength{\tabcolsep}{0.7mm}
\begin{tabular}{lcccccccccccccccc}
\toprule
\multirow{3}{*}{\textbf{Method}} & \multirow{3}{*}{\textbf{Epochs}}
  & \multicolumn{6}{c}{\textbf{COCO $\boldsymbol{1\times}$ schedule}}
  & \multicolumn{6}{c}{\textbf{COCO $\boldsymbol{2\times}$ schedule}}
  & \multicolumn{3}{c}{\textbf{PASCAL VOC}} \\
\cmidrule(lr){3-8} \cmidrule(lr){9-14} \cmidrule(lr){15-17}
& & $\text{AP}^{\text{bb}}$ & $\text{AP}^{\text{bb}}_{50}$ & $\text{AP}^{\text{bb}}_{75}$ & $\text{AP}^{\text{m}}$ & $\text{AP}^{\text{m}}_{50}$ & $\text{AP}^{\text{m}}_{75}$ & $\text{AP}^{\text{bb}}$ & $\text{AP}^{\text{bb}}_{50}$ & $\text{AP}^{\text{bb}}_{75}$ & $\text{AP}^{\text{m}}$ & $\text{AP}^{\text{m}}_{50}$ & $\text{AP}^{\text{m}}_{75}$ & $\mathbf{AP}$ & $\mathbf{AP_{50}}$ & $\mathbf{AP_{75}}$ \\
\midrule
\textit{Supervised} & 200 & 38.2 & 58.2 & 41.2 & 33.3 & 54.7 & 35.2 & 40.0 & 59.9 & 43.1 & 34.7 & 56.5 & 36.9 & 53.5 & 81.3 & 58.8 \\
\textit{Random init} & 200 & 26.4 & 44.0 & 27.8 & 29.3 & 46.9 & 30.8 & 35.6 & 54.6 & 38.2 & 31.4 & 51.5 & 33.5 & 60.2 & 33.1 \\
SimSiam~\citep{chen2020simsiam} & 200 & 39.2 & 59.3 & 42.1 & 34.4 & 56.0 & 36.7 & - & - & - & - & - & - & 57.0 & 82.4 & 63.7 \\
BYOL~\citep{grill2020byol} & 300 & - & - & - & - & - & - & 40.3 & 60.5 & 43.9 & 35.1 & 56.8 & 37.3 & 51.9 & 81.0 & 56.5 \\
SwAV~\citep{caron2020swav} & 800 & 38.4 & 58.6 & 41.3 & 33.8 & 55.2 & 35.9 & - & - & - & - & - & - & 56.1 & 82.6 & 62.7 \\
SimCLR~\citep{chen2020simclr} & 1000 & - & - & - & - & - & - & 40.3 & 60.5 & 43.9 & 35.1 & 56.8 & 37.3 & 56.3 & 81.9 & 62.5 \\
Barlow Twins~\citep{zbontar2021barlowtwins} & 1000 & 39.2 & 59.0 & 42.5 & 34.3 & 56.0 & 36.5 & - & - & - & - & - & - & 56.8 & 82.6 & 63.4 \\
\midrule
\multicolumn{17}{c}{\textit{Detection-specific}} \\
SoCo~\citep{wei2021soco} & 100 & 40.4 & 60.4 & 43.7 & 34.9 & 56.8 & 37.0 & 41.1 & 61.0 & 44.4 & 35.6 & 57.5 & 38.0 & 59.1 & 83.4 & 65.6 \\
InsLoc~\citep{yang2021insloc} & 200 & 39.5 & 59.1 & 42.7 & 34.5 & 56.0 & 36.8 & 41.4 & 60.9 & 45.0 & 35.9 & 57.6 & 38.4 & 57.9 & 82.9 & 64.9 \\
DetCo~\citep{xie2021detco} & 200 & 39.8 & 59.7 & 43.0 & 34.7 & 56.3 & 36.7 & 41.3 & 61.2 & 45.0 & 35.8 & 57.9 & 38.2 & 57.8 & 82.6 & 64.2 \\
ReSim~\citep{xiao2021resim} & 200 & 39.7 & 59.0 & 43.0 & 34.6 & 55.9 & 37.1 & - & - & - & - & - & - & 58.7 & 83.1 & 66.3 \\
\midrule
\multicolumn{17}{c}{\textit{MoCo-based}} \\
MoCo~\citep{he2020moco} & 200 & 38.5 & 58.3 & 41.6 & 33.6 & 54.8 & 35.6 & \underline{40.7} & \underline{60.5} & \underline{44.1} & 35.4 & \underline{57.3} & \underline{37.6} & 55.9 & 81.5 & 62.6 \\
MoCo-v2~\citep{chen2020mocov2} & 200 & 38.9 & 58.4 & 42.0 & 34.2 & 55.2 & 36.5 & \underline{40.7} & \underline{60.5} & \underline{44.1} & \underline{35.6} & \textbf{57.4} & 37.1 & 57.0 & 82.4 & 63.6 \\
MoCHI~\citep{kalantidis2020mochi} & 200 & \underline{39.2} & \underline{58.9} & \underline{42.4} & \underline{34.3} & \underline{55.5} & \underline{36.6} & - & - & - & - & - & - & \underline{57.5} & \underline{82.7} & \underline{64.4} \\
SynCo (ours) & 200 & \cellcolor{pearDarker!08}\textbf{39.9} & \cellcolor{pearDarker!08}\textbf{59.6} & \cellcolor{pearDarker!08}\textbf{43.3} & \cellcolor{pearDarker!08}\textbf{34.9} & \cellcolor{pearDarker!08}\textbf{56.5} & \cellcolor{pearDarker!08}\textbf{36.9} & \cellcolor{pearDarker!08}\textbf{41.0} & \cellcolor{pearDarker!08}\textbf{60.6} & \cellcolor{pearDarker!08}\textbf{44.8} & \cellcolor{pearDarker!08}\textbf{35.7} & \cellcolor{pearDarker!08}\textbf{57.4} & \cellcolor{pearDarker!08}\textbf{38.1} & \cellcolor{pearDarker!08}\textbf{57.7} & \cellcolor{pearDarker!08}\textbf{82.9} & \cellcolor{pearDarker!08}\textbf{64.6} \\
\bottomrule
\end{tabular}
\end{table*}

We transfer the 200-epoch SynCo representation to detection (\Cref{tab:detection_coco_voc}). On PASCAL VOC, SynCo outperforms all MoCo-based methods including MoCHi (\textbf{+0.2} AP) and significantly outperforms the supervised baseline (\textbf{+4.2} AP). On the more challenging COCO, with the 1$\times$ schedule, it improves consistently over the supervised baseline (AP\(^{bb}\) \textbf{+1.7}, AP\(^{m}\) \textbf{+1.6}) and MoCo-v2 (AP\(^{bb}\) \textbf{+1.0}, AP\(^{m}\) \textbf{+0.7}), remaining competitive with detection-specific methods such as DetCo \cite{xie2021detco} and InsLoc \cite{yang2021insloc}.

\section{Discussion}
\label{sec:discussion}

This section examines how synthetic negatives affect proxy task difficulty and shape representation space utilization. We support our analysis with proxy task accuracy curves, alignment and uniformity metrics, and class concentration ratios across MoCo-based methods.

\begin{figure}[!t]
\centering
\includegraphics[width=0.9\linewidth]{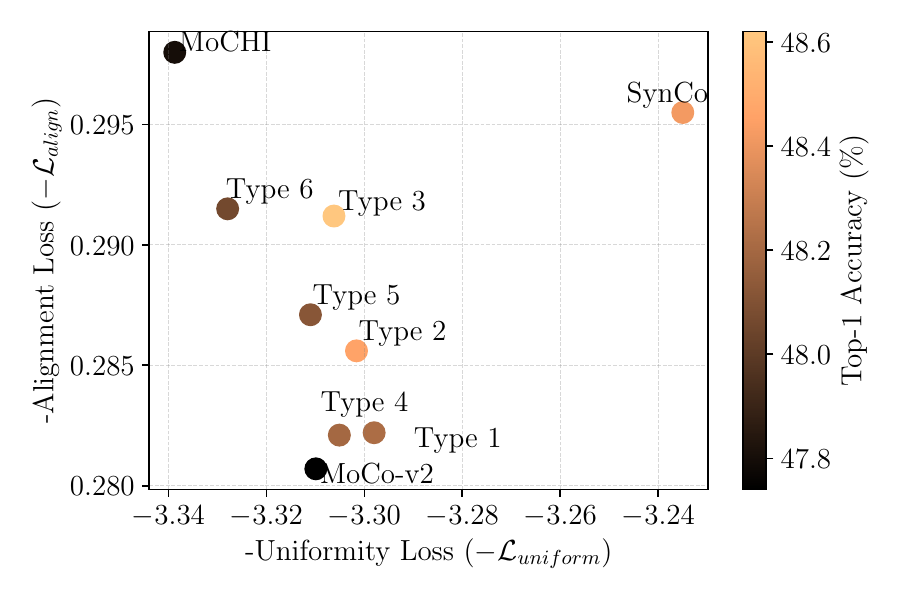}
\caption{\textbf{Alignment and uniformity.} The x- and y-axis represent $-\mathcal{L}_{\text{uniform}}$ and $-\mathcal{L}_{\text{align}}$, respectively. \textbf{Best}: upper-right corner.}
\label{fig:align_uniform}
\end{figure}

\begin{figure}[!t]
\centering
\includegraphics[width=0.9\linewidth]{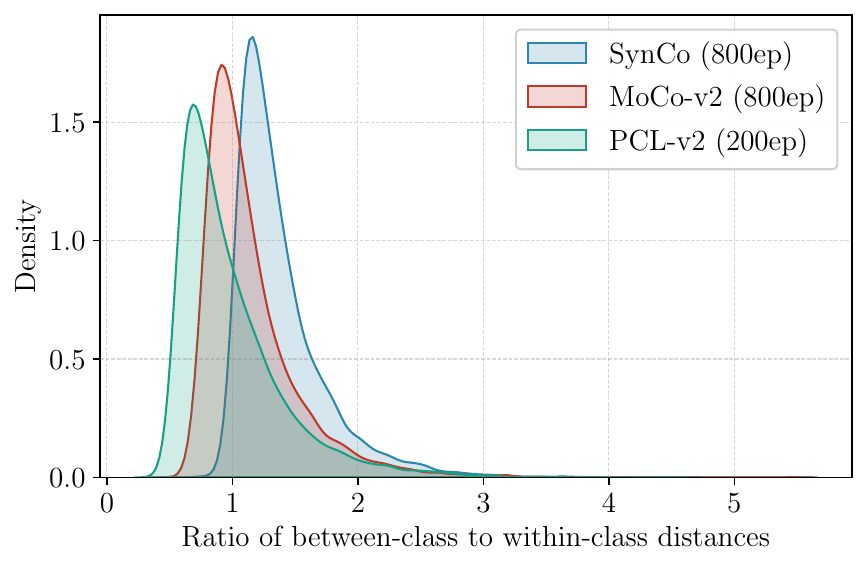}
\caption{\textbf{Distribution of the ratio between inter- and intra-class distances.} Higher values indicate better class separation.}
\label{fig:class_concentration}
\end{figure}

\subsection{Is the Proxy Task More Difficult?}
\label{subsec:proxy_task}

We observe that incorporating synthetic negatives leads to faster learning and improved performance. Each synthetic negative type accelerates learning compared to MoCo-v2, with the full SynCo configuration showing the most significant improvement and lowest final proxy task performance, as shown in \Cref{fig:proxy_task_acc_moco}. This indicates SynCo presents the most challenging proxy task, evidenced by $\max \ell(\mathbf{s}_k^i) > \max \ell (\mathbf{n}_j)$, where $\mathbf{s}_k^i \in S^i$ are synthetic negatives and $\mathbf{n}_j \in {\mathcal{\hat{Q}}}_N$ are original negatives. Through SynCo, we modulate proxy task difficulty via synthetic negatives, pushing the model to learn more robust features.

\subsection{Evaluating the Representation Space}
\label{subsec:alignment_uniformity}

To assess learned representations, we employ alignment and uniformity metrics \cite{tongzhou2020hypersphere}. These metrics provide insights into representation space utilization, with alignment quantifying grouping of similar samples and uniformity measuring spread across the hypersphere. \Cref{fig:align_uniform} presents results for MoCo-based methods. Our findings demonstrate that SynCo significantly improves representation uniformity compared to MoCo-v2 and MoCHi, showing improved utilization of the representation space. Also, incorporating synthetic negatives ($S^1$ to $S^6$) leads to improved alignment. These results suggest that SynCo's approach yields stronger and more well-distributed feature representations.

\subsection{Class Concentration Analysis}
\label{subsec:class_concentration}

To quantify the learned latent space structure, we examine the relationship between within- and between-class distances. \Cref{fig:class_concentration} shows the distribution of ratios between inter- and intra-class $\ell_2$-distances for representations learned by MoCo-based contrastive methods on the ImageNet validation set. A higher mean ratio indicates better concentration within classes while maintaining greater separation between classes, reflecting improved linear separability (aligned with Fisher's linear discriminant analysis \cite{friedman2009elements}). After 800 epochs, SynCo achieves a mean ratio of 1.384, surpassing MoCo-v2 (1.146) and PCL-v2 (0.988).

\subsection{Why Do Synthetic Negatives Help?}
\label{subsec:why_synco_works}

Our results suggest synthetic negatives improve representation learning through two complementary mechanisms. First, they increase proxy task difficulty by replenishing informative negatives, preventing the rapid saturation of standard contrastive learning (\Cref{fig:histogram_probs}). This maintains stronger repulsive gradients and finer discriminative structure. Second, they expand the explored representation space beyond the empirical manifold induced by the memory queue. Geometric strategies promote interpolation and extrapolation between hard examples, stochastic perturbations improve local robustness, and adversarial ones tighten decision boundaries. Together, these effects improve representation uniformity and class separation (\Cref{fig:align_uniform,fig:class_concentration}), yielding features that transfer better to downstream tasks (\Cref{tab:lineval_imagenet_200ep,tab:lineval_imagenet_800ep,tab:semisup_imagenet,tab:detection_coco_voc}). This also explains the cooldown benefit. Once decision boundaries are sufficiently tightened, further synthesis overconstrains the proxy task, so re-anchoring on the data manifold consolidates rather than erodes the learned structure. The same mechanism explains the queue-size dependence in \Cref{sec:ablation}. With a small real pool, synthetic negatives dominate the contrast set and the proxy task becomes too hard for useful gradients, whereas a large queue keeps synthesis complementary. They thus help when they sharpen, rather than overwhelm, the contrast distribution.

% ============================================================
% Ablations
% ============================================================
\section{Ablation Study}
\label{sec:ablation}

We perform ablation studies on ImageNet-100 examining SynCo's hyperparameters, synthetic negative types, and queue size. SynCo is robust to the choice of $\sigma$, $\delta$, and $\eta$ across an order of magnitude of variation (\Cref{tab:ablation1_hyperparams}). For the synthetic negative types, we evaluate each strategy individually, in pairwise and grouped combinations, and all together (\Cref{tab:ablation2_types}). Every individual type improves over the MoCo-v2 baseline, with geometric ($S^1, S^2, S^3$) and gradient-based ($S^5, S^6$) strategies each reaching 48.3\% within their respective groups. Some pairwise combinations interfere (\eg, $S^1, S^3$ and $S^2, S^5$), likely because strategies drawn from the same perturbation family produce redundant negatives that crowd a similar region of the representation space rather than broadening the contrast distribution. This also motivates our count allocation, where the cheaper geometric strategies receive more samples ($N_1=N_2=N_3=256$) than the gradient-based and stochastic ones ($N_4=N_5=N_6=64$). Yet all six together consistently yields the best performance, confirming that the full set provides sufficiently diverse and complementary signals. Regarding queue size (\Cref{tab:ablation3_queue}), SynCo underperforms MoCo-v2 with smaller queues as synthetic negatives become too hard relative to the real pool \cite{kalantidis2020mochi}, but converges and exceeds MoCo-v2 at larger queues. We note that within this strictly controlled setting, where architecture, augmentation strategy, training schedule, and projection dimensions are all held constant across compared methods, consistent improvements across every benchmark configuration provide reliable evidence of method effectiveness independent of orthogonal design choices.

\begin{table}[!tb]
    \centering
    \setlength{\tabcolsep}{1.7mm}
    \caption{\textbf{Ablation on $\boldsymbol{\sigma}$, $\boldsymbol{\delta}$, and $\boldsymbol{\eta}$.} We {\fboxsep 1pt\colorbox{gray!10}{highlight}} the default hyperparameter.}
    \label{tab:ablation1_hyperparams}
    \begin{tabular}{l cc cc cc}
    \toprule
    \multirow{2}{*}{\textbf{Value}} & \multicolumn{2}{c}{$\boldsymbol{\sigma}$} & \multicolumn{2}{c}{$\boldsymbol{\delta}$} & \multicolumn{2}{c}{$\boldsymbol{\eta}$} \\
    \cmidrule(lr){2-3} \cmidrule(lr){4-5} \cmidrule(lr){6-7}
    & Top-1 & Top-5 & Top-1 & Top-5 & Top-1 & Top-5 \\
    \midrule
    \cellcolor{gray!10}
    0.01 & 48.20 & 74.26 & 48.12 & 74.46 & 48.06 & 74.00 \\
    0.05 & 48.36 & 73.84 & 48.88 & 74.72 & 47.16 & 74.14 \\
    0.10 & 47.62 & 73.50 & 48.04 & 73.72 & 47.76 & 74.06 \\
    \bottomrule
    \end{tabular}
\end{table}

\begin{table}[!tb]
\centering
\setlength{\tabcolsep}{1.1mm}
\caption{\textbf{Ablation on queue size $\boldsymbol{K}$.} We {\fboxsep 1pt\colorbox{gray!10}{highlight}} the default hyperparameter.}
\label{tab:ablation3_queue}
\begin{tabular}{lcccccc}
\toprule
\multirow{2}{*}{\textbf{Method}} & \multicolumn{6}{c}{\textbf{Queue size $K$}} \\
& 4k & 8k & 16k & 32k & \cellcolor{gray!10}65k & 131k \\
\midrule
MoCo-v2~\citep{chen2020mocov2} & 50.10 & 50.50 & 49.32 & 48.02 & 47.74 & 47.60 \\
SynCo (ours) & 48.30 & 48.50 & 49.40 & 48.08 & 48.42 & 48.50 \\
\bottomrule
\end{tabular}
\end{table}

\begin{table}[!tb]
    \centering
    \caption{\textbf{Ablation on synthetic negative types.} We convert each configuration into a binary selection over six synthetic negative strategies ($S^1$ to $S^6$). We {\fboxsep 1pt\colorbox{gray!10}{highlight}} the default hyperparameter.}
    \label{tab:ablation2_types}
    \begin{tabular}{cccccccc}
    \toprule
    $\boldsymbol{S^1}$ & $\boldsymbol{S^2}$ & $\boldsymbol{S^3}$ & $\boldsymbol{S^4}$ & $\boldsymbol{S^5}$ & $\boldsymbol{S^6}$ & \textbf{Top-1} & \textbf{Top-5} \\
    \midrule
    \ding{55} & \ding{55} & \ding{55} & \ding{55} & \ding{55} & \ding{55} & 47.7 & 73.9 \\
    \cellcolor{gray!10}\ding{51} & \ding{55} & \ding{55} & \ding{55} & \ding{55} & \ding{55} & 48.2 & 73.9 \\
    \ding{55} & \cellcolor{gray!10}\ding{51} & \ding{55} & \ding{55} & \ding{55} & \ding{55} & 48.2 & 74.1 \\
    \ding{55} & \ding{55} & \cellcolor{gray!10}\ding{51} & \ding{55} & \ding{55} & \ding{55} & 48.2 & 73.8 \\
    \ding{55} & \ding{55} & \ding{55} & \cellcolor{gray!10}\ding{51} & \ding{55} & \ding{55} & 48.2 & 74.2 \\
    \ding{55} & \ding{55} & \ding{55} & \ding{55} & \cellcolor{gray!10}\ding{51} & \ding{55} & 48.1 & 74.4 \\
    \ding{55} & \ding{55} & \ding{55} & \ding{55} & \ding{55} & \cellcolor{gray!10}\ding{51} & 48.0 & 74.0 \\
    \cellcolor{gray!10}\ding{51} & \cellcolor{gray!10}\ding{51} & \ding{55} & \ding{55} & \ding{55} & \ding{55} & 48.1 & 73.9 \\
    \cellcolor{gray!10}\ding{51} & \ding{55} & \cellcolor{gray!10}\ding{51} & \ding{55} & \ding{55} & \ding{55} & 46.8 & 72.9 \\
    \ding{55} & \cellcolor{gray!10}\ding{51} & \cellcolor{gray!10}\ding{51} & \ding{55} & \ding{55} & \ding{55} & 48.2 & 74.1 \\
    \ding{55} & \cellcolor{gray!10}\ding{51} & \ding{55} & \ding{55} & \cellcolor{gray!10}\ding{51} & \ding{55} & 46.3 & 74.1 \\
    \cellcolor{gray!10}\ding{51} & \cellcolor{gray!10}\ding{51} & \cellcolor{gray!10}\ding{51} & \ding{55} & \ding{55} & \ding{55} & 48.3 & 74.2 \\
    \ding{55} & \ding{55} & \ding{55} & \cellcolor{gray!10}\ding{51} & \cellcolor{gray!10}\ding{51} & \cellcolor{gray!10}\ding{51} & 48.3 & 74.1 \\
    \midrule
    \cellcolor{gray!10}\ding{51} & \cellcolor{gray!10}\ding{51} & \cellcolor{gray!10}\ding{51} & \cellcolor{gray!10}\ding{51} & \cellcolor{gray!10}\ding{51} & \cellcolor{gray!10}\ding{51} & \textbf{48.4} & \textbf{74.5} \\
    \bottomrule
    \end{tabular}
\end{table}

\section{Conclusion}
\label{sec:conclusion}

By generating diverse synthetic hard negatives directly in the representation space using six complementary strategies, SynCo improves representation quality with negligible computational overhead. Operating in the representation space makes synthetic negative generation both cheap and practical: all strategies run on cached queue embeddings, with negligible additional cost. We demonstrate that SynCo consistently improves over MoCo-v2~\cite{chen2020mocov2} across pretraining durations and, unlike MoCHi~\cite{kalantidis2020mochi}, continues to benefit from extended pretraining schedules. More broadly, our results indicate that representation-space synthesis is a promising and underexplored direction for enriching the negative distribution in any InfoNCE-based contrastive framework.

\subsection*{Scope and Fair Comparison} 

We deliberately focus on MoCo-based comparisons to ensure fair evaluation under identical training conditions. Recent self-distillation methods like DINO \cite{oquab2023dinov2,simeoni2025dinov3} or iBOT \cite{zhou2022ibot} operate in fundamentally different paradigms, \ie, they use teacher-student architectures, multiple crops, and train on billions of samples with larger computational budgets. Comparing SynCo to these methods would conflate the benefits of synthetic negatives with differences in architecture, data scale, and training. Our contribution specifically targets improving contrastive learning through diverse representation space augmentation. Within this controlled scope, where all confounding factors are held constant, the consistent improvements across every downstream task provide clear evidence validating our approach (\Cref{tab:lineval_imagenet_200ep,tab:lineval_imagenet_800ep,tab:semisup_imagenet,tab:detection_coco_voc}).

% ============================================================
% Acknowledgments
% ============================================================
\section*{Acknowledgments}
 
We acknowledge the computational resources and support provided by the Imperial College Research Computing Service (\url{http://doi.org/10.14469/hpc/2232}), which enabled our experiments.

%%%%%%%%% REFERENCES
{
    \small
    \bibliographystyle{ieeenat_fullname}
    \bibliography{main}

@misc{jahanian2022generative,
	title        = {Generative Models as a Data Source for Multiview Representation Learning},
	author       = {Ali Jahanian and Xavier Puig and Yonglong Tian and Phillip Isola},
	year         = 2022,
	url          = {https://arxiv.org/abs/2106.05258},
	eprint       = {2106.05258},
	archiveprefix = {arXiv},
	primaryclass = {cs.CV}
}

@misc{lin2024accelerating,
	title        = {Accelerating Augmentation Invariance Pretraining},
	author       = {Jinhong Lin and Cheng-En Wu and Yibing Wei and Pedro Morgado},
	year         = 2024,
	url          = {https://arxiv.org/abs/2410.22364},
	eprint       = {2410.22364},
	archiveprefix = {arXiv},
	primaryclass = {cs.CV}
}

@misc{simeoni2025dinov3,
	title        = {DINOv3},
	author       = {Oriane Siméoni and Huy V. Vo and Maximilian Seitzer and Federico Baldassarre and Maxime Oquab and Cijo Jose and Vasil Khalidov and Marc Szafraniec and Seungeun Yi and Michaël Ramamonjisoa and Francisco Massa and Daniel Haziza and Luca Wehrstedt and Jianyuan Wang and Timothée Darcet and Théo Moutakanni and Leonel Sentana and Claire Roberts and Andrea Vedaldi and Jamie Tolan and John Brandt and Camille Couprie and Julien Mairal and Hervé Jégou and Patrick Labatut and Piotr Bojanowski},
	year         = 2025,
	url          = {https://arxiv.org/abs/2508.10104},
	eprint       = {2508.10104},
	archiveprefix = {arXiv},
	primaryclass = {cs.CV}
}

@misc{wei2021soco,
	title        = {Aligning Pretraining for Detection via Object-Level Contrastive Learning},
	author       = {Fangyun Wei and Yue Gao and Zhirong Wu and Han Hu and Stephen Lin},
	year         = 2021,
	url          = {https://arxiv.org/abs/2106.02637},
	eprint       = {2106.02637},
	archiveprefix = {arXiv},
	primaryclass = {cs.CV}
}

@misc{xie2021detco,
	title        = {DetCo: Unsupervised Contrastive Learning for Object Detection},
	author       = {Enze Xie and Jian Ding and Wenhai Wang and Xiaohang Zhan and Hang Xu and Peize Sun and Zhenguo Li and Ping Luo},
	year         = 2021,
	url          = {https://arxiv.org/abs/2102.04803},
	eprint       = {2102.04803},
	archiveprefix = {arXiv},
	primaryclass = {cs.CV}
}

@misc{yang2021insloc,
	title        = {Instance Localization for Self-supervised Detection Pretraining},
	author       = {Ceyuan Yang and Zhirong Wu and Bolei Zhou and Stephen Lin},
	year         = 2021,
	url          = {https://arxiv.org/abs/2102.08318},
	eprint       = {2102.08318},
	archiveprefix = {arXiv},
	primaryclass = {cs.CV}
}

@misc{rojasgomez2024sassl,
	title        = {SASSL: Enhancing Self-Supervised Learning via Neural Style Transfer},
	author       = {Renan A. Rojas-Gomez and Karan Singhal and Ali Etemad and Alex Bijamov and Warren R. Morningstar and Philip Andrew Mansfield},
	year         = 2024,
	url          = {https://arxiv.org/abs/2312.01187},
	eprint       = {2312.01187},
	archiveprefix = {arXiv},
	primaryclass = {cs.CV}
}

@book{friedman2009elements,
	title        = {The Elements of Statistical Learning: Data Mining, Inference, and Prediction},
	author       = {Friedman, Jerome and Hastie, Trevor and Tibshirani, Robert},
	year         = 2009,
	publisher    = {Springer},
	address      = {New York},
	series       = {Springer Series in Statistics},
	doi          = {10.1007/978-0-387-84858-7},
	isbn         = {978-0-387-84857-0},
	edition      = {Second}
}

@article{deng2009imagenet,
	title        = {ImageNet: A large-scale hierarchical image database},
	author       = {Jia Deng and Wei Dong and Richard Socher and Li-Jia Li and K. Li and Li Fei-Fei},
	year         = 2009,
	journal      = {2009 IEEE Conference on Computer Vision and Pattern Recognition},
	pages        = {248--255},
	url          = {https://api.semanticscholar.org/CorpusID:57246310}
}

@article{everingham2009voc,
	title        = {The Pascal Visual Object Classes ({VOC}) Challenge},
	author       = {Mark Everingham and Luc Van Gool and Christopher K. I. Williams and John Winn and Andrew Zisserman},
	year         = 2009,
	month        = sep,
	journal      = {International Journal of Computer Vision},
	publisher    = {Springer Science and Business Media {LLC}},
	volume       = 88,
	number       = 2,
	pages        = {303--338},
	doi          = {10.1007/s11263-009-0275-4},
	url          = {https://doi.org/10.1007/s11263-009-0275-4}
}

@misc{li2015coco,
	title        = {Microsoft COCO: Common Objects in Context},
	author       = {Tsung-Yi Lin and Michael Maire and Serge Belongie and Lubomir Bourdev and Ross Girshick and James Hays and Pietro Perona and Deva Ramanan and C. Lawrence Zitnick and Piotr Dollár},
	year         = 2015,
	eprint       = {1405.0312},
	archiveprefix = {arXiv},
	primaryclass = {cs.CV}
}

@misc{wu2018instdis,
	title        = {Unsupervised Feature Learning via Non-Parametric Instance-level Discrimination},
	author       = {Zhirong Wu and Yuanjun Xiong and Stella Yu and Dahua Lin},
	year         = 2018,
	eprint       = {1805.01978},
	archiveprefix = {arXiv},
	primaryclass = {cs.CV}
}

@misc{hjelm2019dim,
	title        = {Learning deep representations by mutual information estimation and maximization},
	author       = {R Devon Hjelm and Alex Fedorov and Samuel Lavoie-Marchildon and Karan Grewal and Phil Bachman and Adam Trischler and Yoshua Bengio},
	year         = 2019,
	eprint       = {1808.06670},
	archiveprefix = {arXiv},
	primaryclass = {stat.ML}
}

@inproceedings{mehrabi21a,
	title        = {Fundamental Tradeoffs in Distributionally Adversarial Training},
	author       = {Mehrabi, Mohammad and Javanmard, Adel and Rossi, Ryan A. and Rao, Anup and Mai, Tung},
	year         = 2021,
	month        = {18--24 Jul},
	booktitle    = {Proceedings of the 38th International Conference on Machine Learning},
	publisher    = {PMLR},
	series       = {Proceedings of Machine Learning Research},
	volume       = 139,
	pages        = {7544--7554},
	url          = {https://proceedings.mlr.press/v139/mehrabi21a.html},
	editor       = {Meila, Marina and Zhang, Tong}
}

@misc{oord2019cpc,
	title        = {Representation Learning with Contrastive Predictive Coding},
	author       = {Aaron van den Oord and Yazhe Li and Oriol Vinyals},
	year         = 2019,
	eprint       = {1807.03748},
	archiveprefix = {arXiv},
	primaryclass = {cs.LG}
}

@misc{misra2019pirl,
	title        = {Self-Supervised Learning of Pretext-Invariant Representations},
	author       = {Ishan Misra and Laurens van der Maaten},
	year         = 2019,
	eprint       = {1912.01991},
	archiveprefix = {arXiv},
	primaryclass = {cs.CV}
}

@misc{zhuang2019la,
	title        = {Local Aggregation for Unsupervised Learning of Visual Embeddings},
	author       = {Chengxu Zhuang and Alex Lin Zhai and Daniel Yamins},
	year         = 2019,
	eprint       = {1903.12355},
	archiveprefix = {arXiv},
	primaryclass = {cs.CV}
}

@misc{bachman2019amdim,
	title        = {Learning Representations by Maximizing Mutual Information Across Views},
	author       = {Philip Bachman and R Devon Hjelm and William Buchwalter},
	year         = 2019,
	eprint       = {1906.00910},
	archiveprefix = {arXiv},
	primaryclass = {cs.LG}
}

@misc{tian2020cmc,
	title        = {Contrastive Multiview Coding},
	author       = {Yonglong Tian and Dilip Krishnan and Phillip Isola},
	year         = 2020,
	eprint       = {1906.05849},
	archiveprefix = {arXiv},
	primaryclass = {cs.CV}
}

@misc{chen2020simclr,
	title        = {A Simple Framework for Contrastive Learning of Visual Representations},
	author       = {Ting Chen and Simon Kornblith and Mohammad Norouzi and Geoffrey Hinton},
	year         = 2020,
	eprint       = {2002.05709},
	archiveprefix = {arXiv},
	primaryclass = {cs.LG}
}

@misc{grill2020byol,
	title        = {Bootstrap your own latent: A new approach to self-supervised Learning},
	author       = {Jean-Bastien Grill and Florian Strub and Florent Altché and Corentin Tallec and Pierre H. Richemond and Elena Buchatskaya and Carl Doersch and Bernardo Avila Pires and Zhaohan Daniel Guo and Mohammad Gheshlaghi Azar and Bilal Piot and Koray Kavukcuoglu and Rémi Munos and Michal Valko},
	year         = 2020,
	eprint       = {2006.07733},
	archiveprefix = {arXiv},
	primaryclass = {cs.LG}
}

@misc{chen2020mocov2,
	title        = {Improved Baselines with Momentum Contrastive Learning},
	author       = {Xinlei Chen and Haoqi Fan and Ross Girshick and Kaiming He},
	year         = 2020,
	eprint       = {2003.04297},
	archiveprefix = {arXiv},
	primaryclass = {cs.CV}
}

@misc{chen2020simsiam,
	title        = {Exploring Simple Siamese Representation Learning},
	author       = {Xinlei Chen and Kaiming He},
	year         = 2020,
	eprint       = {2011.10566},
	archiveprefix = {arXiv},
	primaryclass = {cs.CV}
}

@misc{he2020moco,
	title        = {Momentum Contrast for Unsupervised Visual Representation Learning},
	author       = {Kaiming He and Haoqi Fan and Yuxin Wu and Saining Xie and Ross Girshick},
	year         = 2020,
	eprint       = {1911.05722},
	archiveprefix = {arXiv},
	primaryclass = {cs.CV}
}

@inproceedings{tian2020infomin,
	title        = {What Makes for Good Views for Contrastive Learning?},
	author       = {Tian, Yonglong and Sun, Chen and Poole, Ben and Krishnan, Dilip and Schmid, Cordelia and Isola, Phillip},
	year         = 2020,
	booktitle    = {Advances in Neural Information Processing Systems},
	publisher    = {Curran Associates, Inc.},
	volume       = 33,
	pages        = {6827--6839},
	url          = {https://proceedings.neurips.cc/paper_files/paper/2020/file/4c2e5eaae9152079b9e95845750bb9ab-Paper.pdf},
	editor       = {H. Larochelle and M. Ranzato and R. Hadsell and M.F. Balcan and H. Lin}
}

@inproceedings{caron2020swav,
	title        = {Unsupervised Learning of Visual Features by Contrasting Cluster Assignments},
	author       = {Caron, Mathilde and Misra, Ishan and Mairal, Julien and Goyal, Priya and Bojanowski, Piotr and Joulin, Armand},
	year         = 2020,
	booktitle    = {Advances in Neural Information Processing Systems},
	publisher    = {Curran Associates, Inc.},
	volume       = 33,
	pages        = {9912--9924},
	url          = {https://proceedings.neurips.cc/paper_files/paper/2020/file/70feb62b69f16e0238f741fab228fec2-Paper.pdf},
	editor       = {H. Larochelle and M. Ranzato and R. Hadsell and M.F. Balcan and H. Lin},
	eprint       = {2006.09882},
	archiveprefix = {arXiv},
	primaryclass = {cs.CV}
}

@misc{mitrovic2020relic,
	title        = {Representation Learning via Invariant Causal Mechanisms},
	author       = {Jovana Mitrovic and Brian McWilliams and Jacob Walker and Lars Buesing and Charles Blundell},
	year         = 2020,
	eprint       = {2010.07922},
	archiveprefix = {arXiv},
	primaryclass = {cs.LG}
}

@misc{kalantidis2020mochi,
	title        = {Hard Negative Mixing for Contrastive Learning},
	author       = {Yannis Kalantidis and Mert Bulent Sariyildiz and Noe Pion and Philippe Weinzaepfel and Diane Larlus},
	year         = 2020,
	eprint       = {2010.01028},
	archiveprefix = {arXiv},
	primaryclass = {cs.CV}
}

@inproceedings{hu2021adco,
	title        = {Adco: Adversarial contrast for efficient learning of unsupervised representations from self-trained negative adversaries},
	author       = {Hu, Qianjiang and Wang, Xiao and Hu, Wei and Qi, Guo-Jun},
	year         = 2021,
	booktitle    = {Proceedings of the IEEE/CVF Conference on Computer Vision and Pattern Recognition},
	pages        = {1074--1083}
}

@misc{khosla2021supcon,
	title        = {Supervised Contrastive Learning},
	author       = {Prannay Khosla and Piotr Teterwak and Chen Wang and Aaron Sarna and Yonglong Tian and Phillip Isola and Aaron Maschinot and Ce Liu and Dilip Krishnan},
	year         = 2021,
	eprint       = {2004.11362},
	archiveprefix = {arXiv},
	primaryclass = {cs.LG}
}

@misc{chen2021mocov3,
	title        = {An Empirical Study of Training Self-Supervised Vision Transformers},
	author       = {Xinlei Chen and Saining Xie and Kaiming He},
	year         = 2021,
	eprint       = {2104.02057},
	archiveprefix = {arXiv},
	primaryclass = {cs.CV}
}

@misc{zbontar2021barlowtwins,
	title        = {Barlow Twins: Self-Supervised Learning via Redundancy Reduction},
	author       = {Jure Zbontar and Li Jing and Ishan Misra and Yann LeCun and Stéphane Deny},
	year         = 2021,
	eprint       = {2103.03230},
	archiveprefix = {arXiv},
	primaryclass = {cs.CV}
}

@misc{dwibedi2021nnclr,
	title        = {With a Little Help from My Friends: Nearest-Neighbor Contrastive Learning of Visual Representations},
	author       = {Debidatta Dwibedi and Yusuf Aytar and Jonathan Tompson and Pierre Sermanet and Andrew Zisserman},
	year         = 2021,
	eprint       = {2104.14548},
	archiveprefix = {arXiv},
	primaryclass = {cs.CV}
}

@misc{caron2021dino,
	title        = {Emerging Properties in Self-Supervised Vision Transformers},
	author       = {Mathilde Caron and Hugo Touvron and Ishan Misra and Hervé Jégou and Julien Mairal and Piotr Bojanowski and Armand Joulin},
	year         = 2021,
	eprint       = {2104.14294},
	archiveprefix = {arXiv},
	primaryclass = {cs.CV}
}

@misc{li2021pcl,
	title        = {Prototypical Contrastive Learning of Unsupervised Representations},
	author       = {Junnan Li and Pan Zhou and Caiming Xiong and Steven C. H. Hoi},
	year         = 2021,
	eprint       = {2005.04966},
	archiveprefix = {arXiv},
	primaryclass = {cs.CV}
}

@misc{xiao2021resim,
	title        = {Region Similarity Representation Learning},
	author       = {Tete Xiao and Colorado J Reed and Xiaolong Wang and Kurt Keutzer and Trevor Darrell},
	year         = 2021,
	eprint       = {2103.12902},
	archiveprefix = {arXiv},
	primaryclass = {cs.CV}
}

@misc{zheng2021ressl,
	title        = {ReSSL: Relational Self-Supervised Learning with Weak Augmentation},
	author       = {Mingkai Zheng and Shan You and Fei Wang and Chen Qian and Changshui Zhang and Xiaogang Wang and Chang Xu},
	year         = 2021,
	eprint       = {2107.09282},
	archiveprefix = {arXiv},
	primaryclass = {cs.CV}
}

@misc{bardes2022vicreg,
	title        = {VICReg: Variance-Invariance-Covariance Regularization for Self-Supervised Learning},
	author       = {Adrien Bardes and Jean Ponce and Yann LeCun},
	year         = 2022,
	eprint       = {2105.04906},
	archiveprefix = {arXiv},
	primaryclass = {cs.CV}
}

@misc{yeh2022dcl,
	title        = {Decoupled Contrastive Learning},
	author       = {Chun-Hsiao Yeh and Cheng-Yao Hong and Yen-Chi Hsu and Tyng-Luh Liu and Yubei Chen and Yann LeCun},
	year         = 2022,
	eprint       = {2110.06848},
	archiveprefix = {arXiv},
	primaryclass = {cs.LG}
}

@misc{tomasev2022relicv2,
	title        = {Pushing the limits of self-supervised ResNets: Can we outperform supervised learning without labels on ImageNet?},
	author       = {Nenad Tomasev and Ioana Bica and Brian McWilliams and Lars Buesing and Razvan Pascanu and Charles Blundell and Jovana Mitrovic},
	year         = 2022,
	eprint       = {2201.05119},
	archiveprefix = {arXiv},
	primaryclass = {cs.CV}
}

@misc{zhu2022tico,
	title        = {TiCo: Transformation Invariance and Covariance Contrast for Self-Supervised Visual Representation Learning},
	author       = {Jiachen Zhu and Rafael M. Moraes and Serkan Karakulak and Vlad Sobol and Alfredo Canziani and Yann LeCun},
	year         = 2022,
	eprint       = {2206.10698},
	archiveprefix = {arXiv},
	primaryclass = {cs.CV}
}

@misc{bardes2022vicregl,
	title        = {VICRegL: Self-Supervised Learning of Local Visual Features},
	author       = {Adrien Bardes and Jean Ponce and Yann LeCun},
	year         = 2022,
	eprint       = {2210.01571},
	archiveprefix = {arXiv},
	primaryclass = {cs.CV}
}

@misc{wang2022clsa,
	title        = {Contrastive Learning with Stronger Augmentations},
	author       = {Xiao Wang and Guo-Jun Qi},
	year         = 2022,
	eprint       = {2104.07713},
	archiveprefix = {arXiv},
	primaryclass = {cs.CV}
}

@article{wang2022caco,
	title        = {CaCo: Both Positive and Negative Samples are Directly Learnable via Cooperative-adversarial Contrastive Learning},
	author       = {Wang, Xiao and Huang, Yuhang and Zeng, Dan and Qi, Guo-Jun},
	year         = 2023,
	journal      = {IEEE Transactions on Pattern Analysis and Machine Intelligence}
}

@misc{bandara2023mixedbarlowtwins,
	title        = {Guarding Barlow Twins Against Overfitting with Mixed Samples},
	author       = {Wele Gedara Chaminda Bandara and Celso M. De Melo and Vishal M. Patel},
	year         = 2023,
	eprint       = {2312.02151},
	archiveprefix = {arXiv},
	primaryclass = {cs.CV}
}

@misc{oquab2023dinov2,
	title        = {DINOv2: Learning Robust Visual Features without Supervision},
	author       = {Oquab, Maxime and Darcet, Timothée and Moutakanni, Theo and Vo, Huy V. and Szafraniec, Marc and Khalidov, Vasil and Fernandez, Pierre and Haziza, Daniel and Massa, Francisco and El-Nouby, Alaaeldin and Howes, Russell and Huang, Po-Yao and Xu, Hu and Sharma, Vasu and Li, Shang-Wen and Galuba, Wojciech and Rabbat, Mike and Assran, Mido and Ballas, Nicolas and Synnaeve, Gabriel and Misra, Ishan and Jegou, Herve and Mairal, Julien and Labatut, Patrick and Joulin, Armand and Bojanowski, Piotr},
	year         = 2023,
	journal      = {arXiv:2304.07193}
}

@misc{estepa2023all4one,
	title        = {All4One: Symbiotic Neighbour Contrastive Learning via Self-Attention and Redundancy Reduction},
	author       = {Imanol G. Estepa and Ignacio Sarasúa and Bhalaji Nagarajan and Petia Radeva},
	year         = 2023,
	eprint       = {2303.09417},
	archiveprefix = {arXiv},
	primaryclass = {cs.CV}
}

@misc{he2015resnet,
	title        = {Deep Residual Learning for Image Recognition},
	author       = {Kaiming He and Xiangyu Zhang and Shaoqing Ren and Jian Sun},
	year         = 2015,
	eprint       = {1512.03385},
	archiveprefix = {arXiv},
	primaryclass = {cs.CV}
}

@misc{ren2016fasterrcnn,
	title        = {Faster R-CNN: Towards Real-Time Object Detection with Region Proposal Networks},
	author       = {Shaoqing Ren and Kaiming He and Ross Girshick and Jian Sun},
	year         = 2016,
	eprint       = {1506.01497},
	archiveprefix = {arXiv},
	primaryclass = {cs.CV}
}

@misc{he2018maskrcnn,
	title        = {Mask R-CNN},
	author       = {Kaiming He and Georgia Gkioxari and Piotr Dollár and Ross Girshick},
	year         = 2018,
	eprint       = {1703.06870},
	archiveprefix = {arXiv},
	primaryclass = {cs.CV}
}

@misc{wu2019detectron2,
	title        = {Detectron2},
	author       = {Yuxin Wu and Alexander Kirillov and Francisco Massa and Wan-Yen Lo and Ross Girshick},
	year         = 2019,
	howpublished = {\url{https://github.com/facebookresearch/detectron2}}
}

@misc{cubuk2019randaugment,
	title        = {RandAugment: Practical automated data augmentation with a reduced search space},
	author       = {Ekin D. Cubuk and Barret Zoph and Jonathon Shlens and Quoc V. Le},
	year         = 2019,
	eprint       = {1909.13719},
	archiveprefix = {arXiv},
	primaryclass = {cs.CV}
}

@misc{arora2019theoretical,
	title        = {A Theoretical Analysis of Contrastive Unsupervised Representation Learning},
	author       = {Sanjeev Arora and Hrishikesh Khandeparkar and Mikhail Khodak and Orestis Plevrakis and Nikunj Saunshi},
	year         = 2019,
	eprint       = {1902.09229},
	archiveprefix = {arXiv},
	primaryclass = {cs.LG}
}

@misc{zhou2022ibot,
	title        = {iBOT: Image BERT Pre-Training with Online Tokenizer},
	author       = {Jinghao Zhou and Chen Wei and Huiyu Wang and Wei Shen and Cihang Xie and Alan Yuille and Tao Kong},
	year         = 2022,
	eprint       = {2111.07832},
	archiveprefix = {arXiv},
	primaryclass = {cs.CV}
}

@misc{balestriero2023cookbook,
	title        = {A Cookbook of Self-Supervised Learning},
	author       = {Randall Balestriero and Mark Ibrahim and Vlad Sobal and Ari Morcos and Shashank Shekhar and Tom Goldstein and Florian Bordes and Adrien Bardes and Gregoire Mialon and Yuandong Tian and Avi Schwarzschild and Andrew Gordon Wilson and Jonas Geiping and Quentin Garrido and Pierre Fernandez and Amir Bar and Hamed Pirsiavash and Yann LeCun and Micah Goldblum},
	year         = 2023,
	eprint       = {2304.12210},
	archiveprefix = {arXiv},
	primaryclass = {cs.LG}
}

@article{zhang2018mixup,
	title        = {Mixup: Beyond empirical risk minimization},
	author       = {Zhang, Hongyi and Cisse, Moustapha and Dauphin, Yann N and Lopez-Paz, David},
	year         = 2017,
	journal      = {arXiv preprint arXiv:1710.09412}
}

@inproceedings{yun2019cutmix,
	title        = {Cutmix: Regularization strategy to train strong classifiers with localizable features},
	author       = {Yun, Sangdoo and Han, Dongyoon and Oh, Seong Joon and Chun, Sanghyuk and Choe, Junsuk and Yoo, Youngjoon},
	year         = 2019,
	booktitle    = {Proceedings of the IEEE/CVF International Conference on Computer Vision},
	pages        = {6023--6032}
}

@inproceedings{tongzhou2020hypersphere,
	title        = {Understanding Contrastive Representation Learning through Alignment and Uniformity on the Hypersphere},
	author       = {Wang, Tongzhou and Isola, Phillip},
	year         = 2020,
	booktitle    = {International Conference on Machine Learning},
	pages        = {9929--9939},
	organization = {PMLR}
}

@inproceedings{hadsell2006dimensionality,
	title        = {Dimensionality Reduction by Learning an Invariant Mapping},
	author       = {Hadsell, R. and Chopra, S. and LeCun, Y.},
	year         = 2006,
	booktitle    = {2006 IEEE Computer Society Conference on Computer Vision and Pattern Recognition (CVPR'06)},
	volume       = 2,
	number       = {},
	pages        = {1735--1742},
	doi          = {10.1109/CVPR.2006.100}
}

@inproceedings{iscen2018mining,
	title        = {Mining on Manifolds: Metric Learning without Labels},
	author       = {Iscen, Ahmet and Tolias, Giorgos and Avrithis, Yannis and Chum, Ondrej},
	year         = 2018,
	booktitle    = {Proceedings of the IEEE Conference on Computer Vision and Pattern Recognition},
	pages        = {7842--7851}
}

@inproceedings{mishchuk2017working,
	title        = {Working Hard to Know Your Neighbor's Margins: Local Descriptor Learning Loss},
	author       = {Mishchuk, Anastasiya and Mishkin, Dmytro and Radenovic, Filip and Matas, Jiri},
	year         = 2017,
	booktitle    = {Advances in Neural Information Processing Systems},
	pages        = {4826--4837}
}

@misc{robinson2021contrastivelearninghardnegative,
	title        = {Contrastive Learning with Hard Negative Samples},
	author       = {Joshua Robinson and Ching-Yao Chuang and Suvrit Sra and Stefanie Jegelka},
	year         = 2021,
	url          = {https://arxiv.org/abs/2010.04592},
	eprint       = {2010.04592},
	archiveprefix = {arXiv},
	primaryclass = {cs.LG}
}

@article{jiang2025hardnegative,
	title        = {Hard-Negative Sampling for Contrastive Learning: Optimal Representation Geometry and Neural- vs Dimensional-Collapse},
	author       = {Ruijie Jiang and Thuan Nguyen and Shuchin Aeron and Prakash Ishwar},
	year         = 2025,
	journal      = {Transactions on Machine Learning Research},
	issn         = {2835-8856},
	url          = {https://openreview.net/forum?id=3cnpZ5SIjU},
	note         = {}
}

@misc{yang2024doesnegativesamplingmatter,
	title        = {Does Negative Sampling Matter? A Review with Insights into its Theory and Applications},
	author       = {Zhen Yang and Ming Ding and Tinglin Huang and Yukuo Cen and Junshuai Song and Bin Xu and Yuxiao Dong and Jie Tang},
	year         = 2024,
	url          = {https://arxiv.org/abs/2402.17238},
	eprint       = {2402.17238},
	archiveprefix = {arXiv},
	primaryclass = {cs.LG}
}

@misc{ge2022robustcontrastivelearningusing,
	title        = {Robust Contrastive Learning Using Negative Samples with Diminished Semantics},
	author       = {Songwei Ge and Shlok Mishra and Haohan Wang and Chun-Liang Li and David Jacobs},
	year         = 2022,
	url          = {https://arxiv.org/abs/2110.14189},
	eprint       = {2110.14189},
	archiveprefix = {arXiv},
	primaryclass = {cs.CV}
}
}

%%%%%%%%% END
\end{document}